\documentclass[10pt,twocolumn,letterpaper]{article}

\usepackage{cvpr}
\usepackage{times}
\usepackage{epsfig}
\usepackage{graphicx}
\usepackage{amsmath}
\usepackage{amssymb}
\usepackage{multicol}
\include{pythonlisting}
\makeatletter
\newcommand{\removelatexerror}{\let\@latex@error\@gobble}
\makeatother
\usepackage{algorithm} 
\usepackage{algpseudocode}

\usepackage{graphicx}
\usepackage{multicol}
\graphicspath{{./images/}}

\usepackage[breaklinks=true,bookmarks=false]{hyperref}

\cvprfinalcopy 

\def\cvprPaperID{7037} 

\begin{document}

\title{Imitation Learning for Fashion Style Based on Hierarchical Multimodal Representation}

\author{Shizhu Liu\\
JD.com American Technologies\\
Mountain View, CA, USA\\
{\tt\small shizhu.liu@jd.com}
\and
Shanglin Yang\\
JD.com  American Technologies\\
Mountain View, CA, USA\\
{\tt\small shanglin.yang@jd.com}
\and
Hui Zhou\\
JD.com  American Technologies\\
Mountain View, CA, USA\\
{\tt\small hui.zhou@jd.com}
}

\maketitle

\begin{abstract}
   Fashion is a complex social phenomenon. People follow fashion styles from demonstrations by experts or fashion icons. However, for machine agent, learning to imitate fashion experts from demonstrations can be challenging, especially for complex styles in environments with high-dimensional, multimodal observations. Most existing research regarding fashion outfit composition utilizes supervised learning methods to mimic the behaviors of style icons. These methods suffer from distribution shift: because the agent greedily imitates some given outfit demonstrations, it can drift away from one style to another styles given subtle differences. In this work, we propose an adversarial inverse reinforcement learning formulation to recover reward functions based on hierarchical multimodal representation (HM-AIRL) during the imitation process. The hierarchical joint representation can more comprehensively model the expert composited outfit demonstrations to recover the reward function.  We demonstrate that the proposed HM-AIRL model is able to recover reward functions that are robust to changes in multimodal observations, enabling us to learn policies under significant variation between different styles.
\end{abstract}

\section{Introduction}

Fashion plays important and sophisticated roles in various aspects: social, culture, identity and etc. We can roughly understand a fashion is a type of reaction common to a considerable number of people\cite{Alpher02}. It is common that fashion styles suggested by fashion experts, fashion icons or more popularly now by Key Opinion Leaders (KOLs) from the social media are imitated by people.

On internet, a set of outfits with description is a common way to demonstrate a fashion style. People imitate the fast changing fashion styles from these demonstrations without interaction with the experts. However, learning fashion styles from demonstrated outfits by machine is challenging, as it requires how low-level fashion elements can be mapped to high-level styles.

\begin{figure}[t]
\begin{center}
   \includegraphics[scale=0.12]{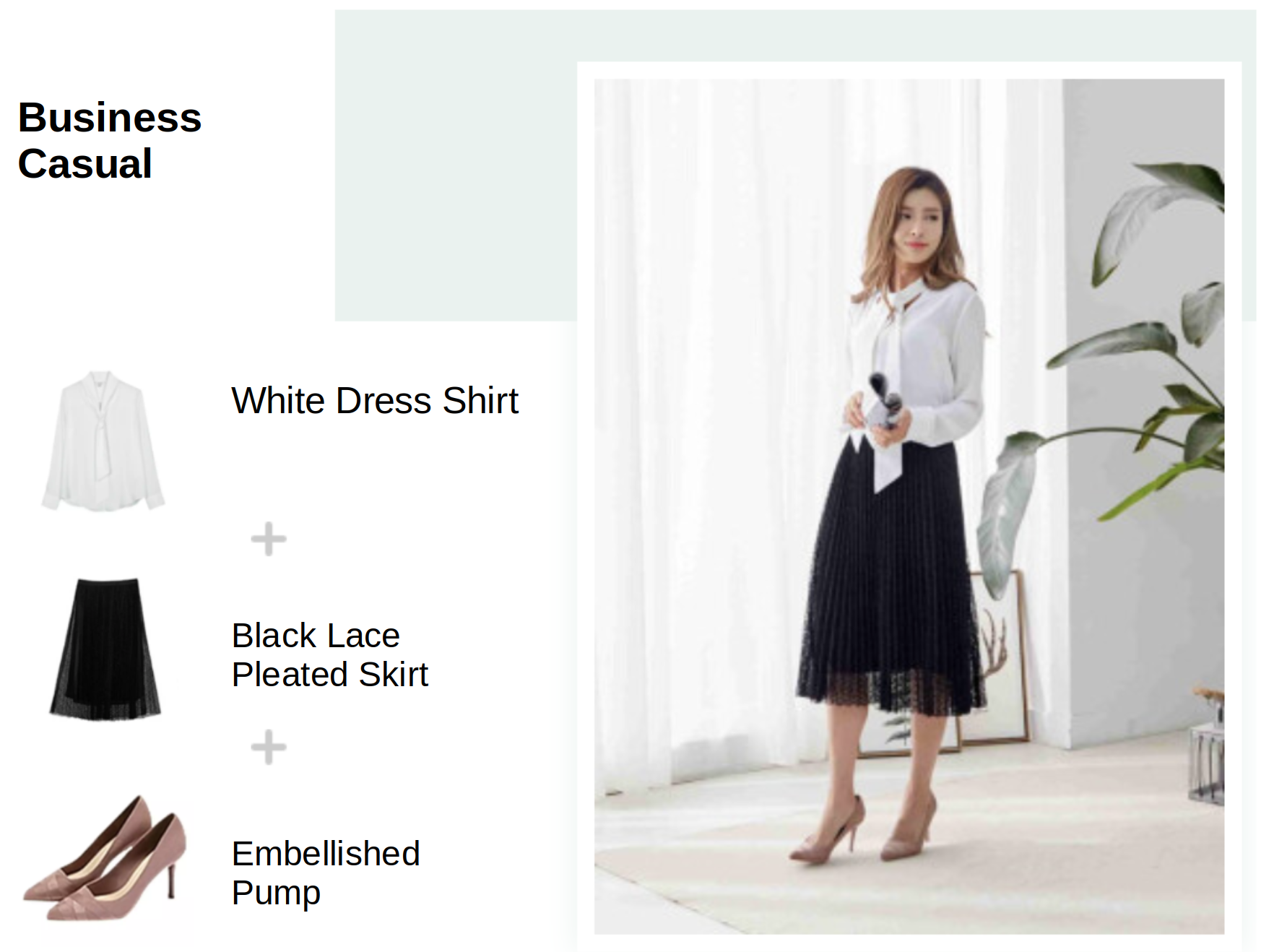}
\end{center}
   \caption{An Outfit Example of "Business Casual" Style}
\label{fig:outfit}
\end{figure}

Fashion styles are composed of important design elements such as color, pattern, material, silhouette, and trim \cite{Sorger_2006_fashion_design}. Each outfit item may demonstrate some design elements with corresponding images and attribute information at low level. High level information for style may be available too: relationship between these items and sometime related cultural meaning behind the suggested style. All the items in an outfit should be consistent with the described style, and be compatible with each other as well. 

Consider the outfit information as input data, learning-based method can be used if we have a good data set with labels of each item and corresponding style. Most existing outfit composition methods follow this supervised learning approach and concentrate on predicting the compatibility between fashion items. However, compatibility is only part of fashion style knowledge. Compatible items cannot guarantee the consistency with the condition style definition. Besides supervised learning methods based on behavioral cloning (BC) suffer from distribution shift: because the agent greedily imitates demonstrated actions, it can drift away from one style to another styles due to subtle differences.

In this paper, we address this fundamental question from two aspects. First, we design a hierarchical multimodal representation to describe complex latent information of the whole outfit structure. Based on this representation, we further propose to use an inverse reinforcement learning method to infer experts' composition reward function and learn the composition value function simultaneously.

Corresponding to the three parts of the outfit demonstration: image, attributes and style description, our outfit encoder network consist of three parts to learn the rich information conveyed in a outfit. At low level, a shared multimodal variational autoencoder is employed to learn the jointly representation from image and attribute information for each item. At high level, we try to cover the the whole outfit style by applying two steps:  matching strategy is applied to extract the relations between pairs of items once the item vectors are generated \cite{Conneau_2017_emnlp}; and a pre-trained BERT model \cite{Devlin_2019_bert} is used to encode explanation text as the condition of the whole outfit.

Adversarial inverse reinforcement learning (AIRL)\cite{Fu_2018_LCLR} is used in our algorithm to provide the agent the capacity for simultaneous learning of the reward function and value function, which enables us to both make use of the efficient adversarial formulation and recover a generalizable and robust reward function for both item compatibility and style consistency.

Our key contributions can be summarized as follows:
\begin{itemize}
  \item We propose a learning-based framework for effective fashion imitation. To our best knowledge, our method is the first to address the imitation behavior in the fashion style learning. 
  
  \item 
  We design a hierarchical multimodal neural network that can effectively encode the rich latent information of whole outfit: at low level, it can capture complementary information from multimodal observations; and interleaving factors are covered at high level.
  
  \item We propose an adversarial inverse reinforcement learning method for recovering the style reward function, which can learn more robust reward to avoid the style drift.
  
  
\end{itemize}

For the rest of the paper, we fist discuss the related work in Section 2. We then describe the hierarchical multi-modality fusion model for outfit in Section 3. Next, details of the proposed adversarial training method is introduced. We show our experimental results in Section 5 and give our discussion and future work at the end.  

\section{Related Work}

Computer vision and recommendation techniques have important and rich applications in fashion domain. The majority of research in this domain focus on fashion image attribute recognition and retrieval, fashion item semantic embedded learning, fashion recommendation, visual compatibility learning and outfit composition.

\textbf{Fashion Attribute Recognition and Retrieval}. Clothing attributes provide a useful tool to assess clothing products as mid-level semantic visual concepts. Recently, more deep networks such as DeepFashion\cite{Liu_2016_CVPR} and MTCT\cite{Dong_2017}, were proven to be efficient in attribute recognition on large datasets. Nanoto et al.\cite{Inoue2017MultilabelFI} proposed a multi-label joint learning network to predict cloth attributes from images with minimum human supervision. Several works utilized weakly labeled image-text pairs to discover attributes[\cite{Berg2010AutomaticAD}, \cite{DBLP_eccv}]. Besides, images retrieval and products recommendation tasks can benefit from attributes results. Chen et al.\cite{chen2015dda} focused on solving the problem of describing people based on fine-grained clothing attributes. AMNet\cite{zhao2017mane} conducted a fashion search after changing an attribute. Mix and match \cite{BMVC2015_51} combined a deep network with conditional random fields to explore the compatibility of clothing items and attributes. Wei et al.\cite{hsiao_fashion_iccv_17} considered both global and localized attributes as 'words' to describe cloth. Many works above utilized attributes for image retrieval and recommendation tasks, but the attributes can not address the high
level information of the whole outfit. In our work, we combine multi-modal information as a global feature. The experiments reveal our feature could also be used for missing attributes inference.

\textbf{Fashion Visual-semantic Embedding Learning}. Fashion item representation learning is the fundamental step of all downstream inference work. There are a lot of works trying to investigate this important problem with different network structure and learning methods. The most common approaches tend to be trained as siamese network\cite{Bromley:1993} or using triplet loss \cite{Schultz:2003}, This is extended in Han et al. \cite{han_fashion_lstm_14} by feeding the visual representation from each garment within an outfit into an LSTM in order to jointly reason about the outfit as a whole. Simo-Serra et al.\cite{simo_2016_style} trained a classifier network with ranking as the constraint to extract distinctive feature representation and also used high level feature of the classifier network as embedding of fashion style. Many unsupervised representation learning methods also were proposed to learn latent feature from unlabeled data directly. Most of them utilize Variational Auto-Encoder (VAE)\cite{kingma_2013_vae} and Predictability Minimization (PM) model\cite{schmidhuber_2008_pm} to learn fashion item embeddings. The encoded embeddings usually contain mixed and unexplainable features of original images. There are several approaches which implemented multi-modal embedding methods to reveal novel feature structures (e.g. [\cite{iccv_HalahSG17},\cite{Shih2017CompatibilityFL},\cite{hsiao_fashion_iccv_17}]). These multimodal methods only to map the image text into the same space without considering the deep correlation between the different modals. In this paper, we try to learn the distributed representation for specific fashion styles. We assume that the complementary representation for fashion style should cover both compatibility between items and common feature shared by whole outfit. To reach this goal, our method learn the joint representation for fashion items from image and attributes information at low level, and capture the relationship between items and condition style.

\textbf{Fashion Recommendation and Outfit Composition}. There are a few approaches for fashion items recommendation. In the context of fashion analysis, visual compatibility measures whether clothing items complement one another across visual categories. For example, “sweat pants” are more compatible with “running shoes” than “high-heeled shoes”. Most existing fashion related research work mainly focus on the compatibility between fashion items, and mainly learning on images data only. Iwata et al.\cite{iwata_2011_ijcai} proposed a topic model to recommend ”Tops” for ”Bottoms”. The goal of this work is to compose fashion outfit automatically by building product coordinates from visual features in each fashion item region.  Veit et al.\cite{veit_2015_iccv} built a Siamese Convolutional Neural Network (CNN) architecture to learn clothing matching pair products from the Amazon co-purchase dataset, focusing on the representative problem of learning compatible clothing style. Simo-Serra1 et al. \cite{simo_2015_iccvpr} introduced a Conditional Random Field (CRF) to learn the different outfits formula and types of people. The model is further being used to predict how fashionable a person looks on a particular photograph. Li et al. \cite{li_2017_itm} used multi-modal embeddings as features and the quality scores as the label to train a grading model. Xuemeng et al. \cite{Song2017NeuroStylistNC} propose to model the compatibility between fashion items based on Bayesian personalized ranking (BPR). Han etc\cite{han_fashion_lstm_14} jointly learn compatibility relationships among fashion items and employ a Bi-LSTM model to learn the compatibility relationships among fashion items by modeling an outfit as a sequence.  \cite{capsule_2018_cvpr} used style topic models for compatibility and defined the recommendation as a subset selection problem. Chen et al. \cite{Chen_2019} proposed an encoder-decoder model to generate personalized fashion outfits. Most of works above utilized supervised learning methods to predict the compatibility between fashion items and validated model performance on validation dataset generated with negative sampling techniques. Thus, these methods are inclined to learn the general patterns rather than capture the subtle difference between fashion styles. To address this problem, our method implemented adversarial learning methods to learns more robust composition policy. 

\textbf{Imitation Learning}. Imitation learning techniques aim to mimic human behavior in a given task. An agent, i.e., a learning machine, by learning a mapping between observations and actions, is trained to perform a task from demonstrations \cite{Billard2016IL}. Inverse reinforcement learning (IRL) is a form of imitation learning that accomplishes this by first inferring the expert's reward function and then training a policy to maximize it\cite{ziebart2008maximum} \cite{Fu_2018_LCLR} \cite{Finn_2016_ICML} \cite{Argall_2009_RAS} \cite{abbeel_2004_irl}. Most of imitation learning works are mainly focus on the imitation process in the fields of robotics, adaptive planning, and data-driven animation. In this work, we formalize the outfit composition task as a Markov Decision Process(MDP) and work under the maximum ambiguity causal IRL framework of \cite{ziebart_2010_thesis}, which allow us to cast the reward learning problem as a maximum likelihood problem. Our IRL algorithm is built upon the adversarial IRL architecture proposed in \cite{Finn_2016} and \cite{Fu_2018_LCLR}. A discriminator is trained to distinguish experts' selection, while the agent is trained to "fool" the discriminator into thinking itself is the expert. To our knowledge, this is the first approach that considers the outfit composition as a style imitation learning problem in the fashion domain.

\section{Hierarchical Multimodal Representation for Fashion Outfit}
\subsection{Fusion Representation for Fashion Item}
\begin{figure}[t]
\begin{center}
   \includegraphics[scale=0.6]{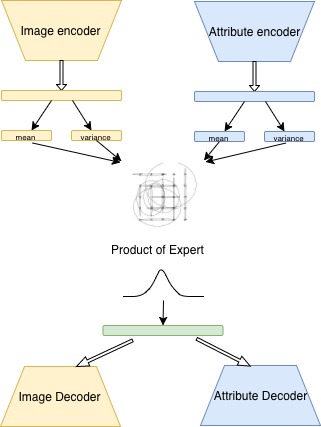}
\end{center}
   \caption{Multimodal Variational Autoencoder Learning on Image and Attributes}
\label{fig:poe}
\end{figure}

For one fashion item, it is obvious that the corresponding image and attribute tags have the complementary information. For instance, people can easily tell the color and pattern of a garment from image, but need to check the attribute tags to know the garment material and the specific functional usage. Learning from diverse modalities with generative approaches has the potential to yield more generalized joint representations. Inspired by the product-of-experts(PoE) inference network \cite{hinton_2006},  assuming the conditional independence among the modalities and joint posterior from multiple modalities is a product of individual posteriors, we utilize a multimodal variational autoencoder (MVAE) \cite{wu_nips_2018} to learn a joint distribution from both image and attribute tags as shown in Figure \ref{fig:poe}.

In the multimodal setting, we assume the $N$ modalities, $x_1, ..., x_N$, are conditionally independent given the common latent variable, $z$. Then, we assume that a generative model of the form $p_\theta(x_1, x_2, ..., x_N, z)=p(z)p_\theta(x_1|z)p_\theta(x_2|z)\cdot\cdot\cdot p_\theta(x_N|x)$. 
If an item is presented by a collection of modalities $X=\{x_i|i^{th} \textrm{modality  present}\}$, then the evidence lower bound(ELBO) becomes:
\begin{equation}
\label{eq:elbo}
\begin{aligned}
    ELBO(X)= \mathbb{E}_{q_\phi(z|X)}[\sum_{\substack{x_ii\in X}}\lambda_i log p_\theta(x_i|z)] \\
    - \beta KL[q_\phi(z|X), p(z)]
\end{aligned}
\end{equation}

MVAE can be trained by simply optimizing the evidence lower bound given in Eqn.\ref{eq:elbo}. The product and quotient distributions are not in general solvable in closed form. However, when $p(z)$ and $\tilde{q}(z|x_i)$ are Gaussian there is a simple analytical solution\cite{cao_2014_poe}: a product of Gaussian experts is itself Gaussian with mean $\mu=(\sum_i\mu_iT_i)(\sum_iT_i)^{-1})$ and covariance $V_i=(\sum_iT_i)^{-1}$, where $\mu_i$, $V_i$ are the parameters of the i-th Gaussian expert, and $T_i=V_i^{-1}$ is the inverse of the covariance.

\subsection{Language-Conditioned Hierarchical Representation for Outfit Style}
Comparing with standard compatibility prediction task, the fashion imitation learning is to learn a reward function that generalizes across different types of styles. While standard supervised methods are typically trained and evaluated without considering style information, we want our language-conditioned reward function to produce correct behavior when generating outfit with given style conditions. 

\begin{figure*}
    \begin{center}
    \includegraphics[scale=0.3]{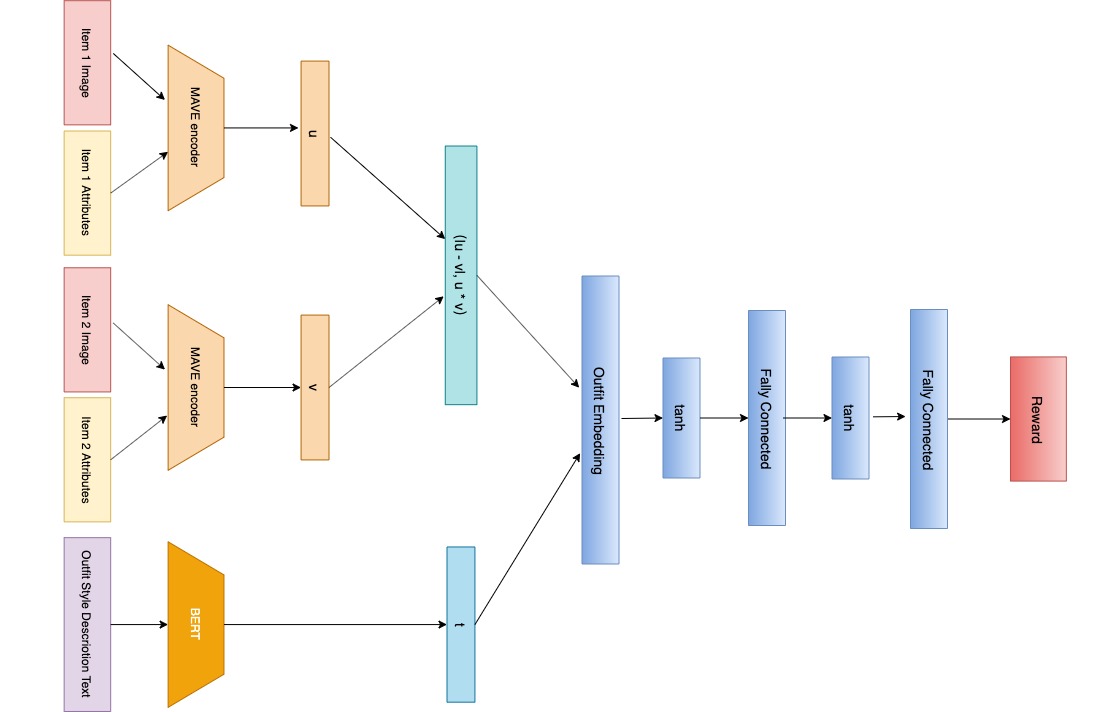}
    \end{center}
   \caption{Language-conditioned Hierarchical Multimodal Network for Style Consistency Reward Learning}
    \label{fig:hierarchical-multimodal-style}
\end{figure*}

Notice that the description text of the outfits usually explains the common salient features such as occasion, season, trending information with which all the items consistent. We encode the explanation language as the condition of the whole outfit structure. As shown in Figure \ref{fig:hierarchical-multimodal-style}, for any pairs of items in an outfit, a pretrained BERT model \cite{Devlin_2019_bert} is used to encode description text $t$, and the MVAE encoder convert two items' image and attributes information to an joint representation $u, v$. The interaction between items is represented as $(|u-v|, u*v)$. The outfit embedding is the concatenation of these two parts, and the outfit style rewards will be learned based on it.

\section{Adversarial Inverse Reinforcement learning for Style Reward Learning}

For the outfit composition task, we formalize the process as follow. Let $I$ denotes the set of all fashion items, $O_i$ denote an outfit, and $x_{i,j} \in I$ denote the items in the outfit $O_i$, so that $O_i = \{x_{i,1},x_{i,2},...,x_{i,|S_i|}\}$. Each item $x_{i,j}$ belong to a limited number of categories $\{C_1,C_2,...,C_N\}$.  The fashion outfit composition process can be formulated as an iterative item selection process, in which at most one item is selected for each category. For example, a user may want to compose an outfit of "UK Smart Casual Style". Then, he/she needs to select one item from four categories: "Shirt", "Jacket", "Pant" and "Shoes" respectively. 

This process can be described with the the maximum causal entropy IRL framework \cite{ziebart_2010_thesis}, which considers an entropy-regularized Markov decision process (MDP), defined by the tuple $(S, A, T, r, \gamma, \rho_0)$. $S,A$ are the state and action spaces, respectively, $\gamma \in (0,1)$ is the discount factor, the dynamics and transition distribution $T(s'|a,s)$, the initial state distribution $\rho_0 (s)$, and the reward function $r(s,a)$ is unknown in the standard reinforcement learning setup and can only be queried through interaction with the MDP. Specifically, $S$ can be presented with the selected fashion items and $A$ refer to the item selection actions during the process. The reward function $r(s,a)$ indicate the compatibility and style consistency between fashion items and the described style.

Because the reward function $r(s,a)$ is unknown, we assume the experts' demonstration outfits are composited with an optimal policy $\pi^{*}(a|s)$. Inverse reinforcement learning instead seeks inferring the reward function $r(s,a)$ given a set of demonstrations $D=\{\tau_1,...,\tau_N\}$. Moreover, the dynamics of composition process is known. Instead of using full trajectories, we could focus on the single state and action case. The entire training procedure is detailed in Algorithm 1. During the training process, our algorithm alternate between training a discriminator to classify the expert selection from outfit generated by current policy, and update the policy to confuse the discriminator \cite{ho_2016_gail} \cite{Finn_2016} \cite{Fu_2018_LCLR}. The discriminator is trained with the form:
\begin{equation}
    D_{\theta , \phi}(s,a,s')=\frac{exp\{ f_{\theta,\phi}(s,a,s')\}}{exp\{ f_{\theta,\phi}(s,a,s')\} + \pi (a|s)^{'}}
\end{equation}
where $f_{\theta,\phi}$ is restricted to a reward approximator $g_{\theta}$ and a shaping term $h_{\phi}$ as
\begin{equation}
    f_{\theta,\phi}(s,a,s') = g_{\theta}(s,a) + \gamma h_{\phi}(s') - h_{\phi}(s)
\end{equation}

Suppose we are given an expert policy $\pi_E$ that we wish to rationalize with IRL. $r^*$ is the true reward function. The $f^*$ is the advantage function need to be recoverd. $h$ recovers the optimal value function $V^*$, which servers as the reward shaping term: 
\begin{equation}
f^{*}(s,a,s^{'}) = r^{*}(s) + \gamma V^*(s^{'}) - V^{*}(s) = A^{*}(s,a)
\end{equation}

\begin{algorithm*}
	\caption{Language-Conditioned Style Reward learning} 
	\begin{algorithmic}[1]
	    \State Obtain expert outfit demonstrations $\tau_E$
	    \State Initialize policy $\pi$ and discriminator $D_{\theta,\phi}$
	    \For {style $t_k \in \{t_1,t_s,...t_N\}$}
		\For {$iteration=1,2,\ldots$}
		    \State Composite outfits $\tau_i = (s_0,a_0,...,s_T,a_T)$ with given style $t_k$ by executing $\pi$.
			\State Train $D_{\theta,\phi}$ via binary logistic regression to classify demonstrations $\tau_E$ from generated outfits $\tau_i$ 
			\State Update reward $r_{\theta,\phi}(s,a,s')\leftarrow log D_{\theta, \phi}(s,a,s') - log(1-D_{\theta,\phi}(s,a,s'))$
			\State Update $\pi$ with respect to $r_{\theta,\phi}$ using policy optimization method.
		\EndFor
		\EndFor
	\end{algorithmic} 
\end{algorithm*}

\section{Experimental Results}

In this section, we introduce the data collection we used for style imitation learning, the evaluation of the composition agent, and some further analysis.

\subsection{Dataset}

Different from fashion datasets collected from Polyvore \cite{han_fashion_lstm_14} \cite{song_2017_nnc} or Lookastic \cite{yang_2019_ifm} that are suitable for data mining, the dataset with more complete fashion style information works better for fashion imitation. Namely, we like the dataset has an explanation text to describe the style of each outfit, and an optional list of every demonstrated item for the given style. This is natural for the human being imitation and many fashion e-commerce websites use this way to demonstrate fashion items on their platforms. We specifically found a website named Chuanda on wqs.jd.com that is suitable for our need. On Chuanda website, all the outfits are curated by fashion experts and every item is mapped to an identical product for sale. 

We collect a fashion style dataset from Chuanda containing 3,557 outfits covering 67 basic fashion styles. In this dataset, each outfit is composed of up to three items, and a short description about the outfit style. Moreover, for each given outfit, there is averagely around 21.04 other candidate items suggested by fashion experts. Every item in this dataset is mapped to an identical sale item on JD.com website. This is convenient and useful as we can collect the product name, images, and attributes for the corresponding fashion item. In another word, in the dataset all these fashion items are labeled with 1,879 distinct fashion related attributes that belong to 5 types: Gender, Season, Style, Material, and Function. More statistics of this dataset is shown in Table \ref{tab:dataset}.

\begin{table}
\begin{center}
\begin{tabular}{|c|c|c|c|c|}
\hline
Outfits & Items & Attributes & Basic styles & Avg Opts \\
\hline
3557 & 18627 & 1879 & 67 & 21.04 \\
\hline
\hline
Top & Bottom & Skirt & Shoes & Coat\\
\hline
6486 & 5134 & 2082 & 2237 & 1731 \\
\hline
\end{tabular}
\end{center}
\caption{Chuanda fashion style dataset statistics}
\label{tab:dataset}
\end{table}

\subsection{Evaluation}
We perform two different evaluations for our proposed learning framework. First, we evaluate the effectiveness of the learned multimodal representation by predicting missing attributes of the given fashion item. Second, we measure style consistency by computing the similarity of composited outfits with the recommendation list provided by fashion experts. 

\begin{figure*}
\begin{center}
   \includegraphics[scale=0.7]{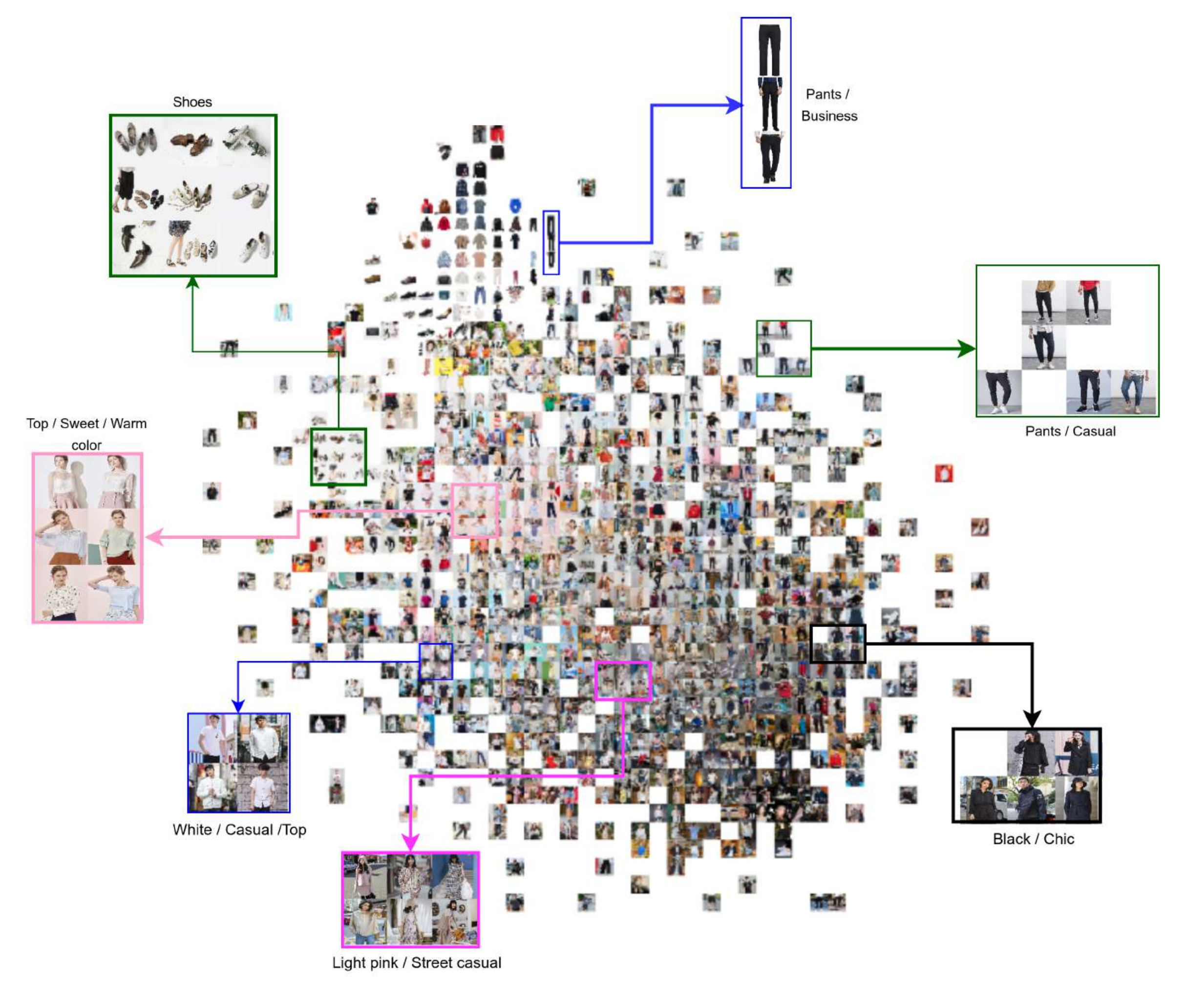}
\end{center}
   \caption{Visualization of the fashion item representation using t-SNE \cite{vanDerMaaten2008}}
\label{fig:clustering}
\end{figure*}

\textbf{Missing Attribute Imputation}. Fashion item representation is critical for the downstream style imitation learning.To verify if our method can learn more complementary information, we conduct the missing attribute imputation task to evaluate the effectiveness of the MVAE. On Chuanda dataset, we simulate incomplete supervision by randomly reserving a fraction of the dataset as multi-modal examples. We examine the effect of supervision on the attribute prediction task $p(x_2|x_1)$, e.g. predict the correct attribute label $x_2$ from an image $x_1$. For the MVAE, the total number of examples shown to the model is always fixed, only the proportion of complete bi-modal example is varied. Five important types of attributes (Gender, Season, Style, Material, and Function) are masked first in the input and then predicted with MVAE decoder, the evaluation results are provided in Table \ref{tab:attributes}.  We also perform a qualitative analysis of the items representation generated from MVAE and visualize the features space in Figure \ref{fig:clustering} using t-SNE \cite{vanDerMaaten2008}. Our representation display robustness to background variance and items share similar style and visual appearance can be clustered together. For example, items of "casual white tops" and "business style pants" are very close in the space regardless of the background noise.  

\begin{table}
\begin{center}
\begin{tabular}{|c|c|c|}
\hline
Attribute Type & \# & Accuracy \\
\hline
Gender & 4529 & 89.26\% \\
\hline
Season & 6928 & 64.3\%\\
\hline
Style & 8143 & 57.71\%\\
\hline
Material & 7081 & 65.6\%\\
\hline
Function & 6957 & 73.7\%\\
\hline
\end{tabular}
\end{center}
\caption{Missing Attribute Prediction}
\label{tab:attributes}
\end{table}

\begin{figure}[t]
\begin{center}
   \includegraphics[scale=0.5]{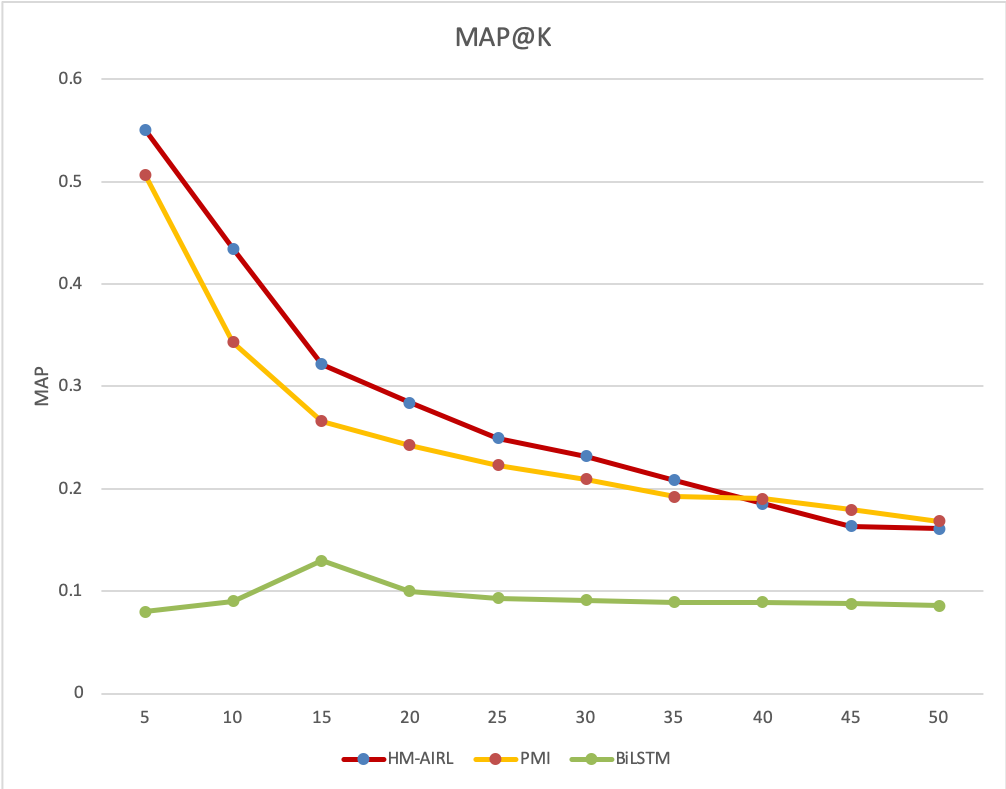}
\end{center}
   \caption{Performance of different methods with respect to MAP at different numbers of top matching items}
\label{fig:map}
\end{figure}

\begin{figure*}
\begin{center}
   \includegraphics[scale=0.6]{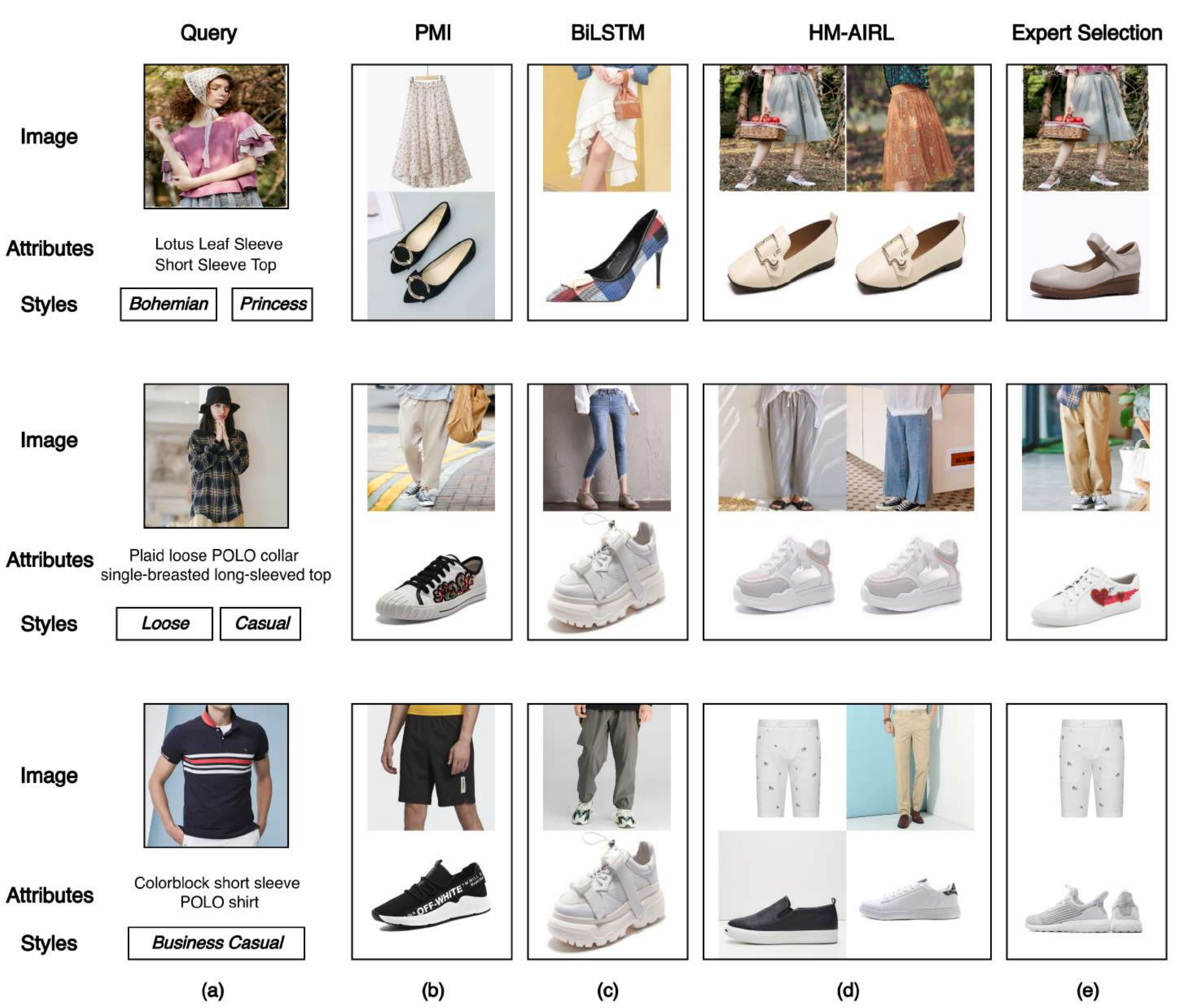}
\end{center}
   \caption{Outfit composition results from different methods with the same condition style and the same querying top item. Three different condition styles and the querying items are listed in column (a). Experts' selections in the demonstration outfits are shown in column (e). Outfits generated by PMI and BiLSTM methods are listed in column (b) and column (c), respectively. Column (d) shows top 2 outfits by the proposed method.}
\label{fig:outfit-results}
\end{figure*}

\textbf{Style Consistency}. In every Chuanda outfit, each item have a list of optional alternatives, which are also consistent with the given style. For condition style and query top, we adopted the common strategy \cite{koren2018} that feeds selected top item and conditional style description as query, and randomly selected K bottoms as the candidates. The item in the experts' demonstration and optional alternatives are positive candidates. Thus, we can evaluate the effectiveness of the imitation learning by measuring the average position of the consistent item in the ranking list with the mean average precision (MAP) metric. We have totally 1000 unique tops and styles in test set.
We compared out method (HM-AIRL) with two methods: feature based pointwise mutual information (PMI) ranking algorithm and Han et al's BiLSTM based method\cite{han_fashion_lstm_14}. Pointwise mutual information is s a measure of association and is used for finding collocations and associations between items. In Chuanda dataset, all the items are labeled with 1,879 attributes. We pre-calculate the the PMI scores between any pair of attribute, and rank the candidate items with the sum of attribute PMI scores. For the Bi-LSTM method, we follow the setting in \cite{han_fashion_lstm_14} and retrain the network on Chuanda dataset. 

The performance of three methods at different number of top k candidate items is shown in Figure \ref{fig:map}. HM-AIRL method get the highest MAP from top 5 to 35 ranking results. We also notice that many items selected by HM-AIRL but not in experts optional list are also compatible with the query top and consistent with the conditional style, which is reasonable in the real application. Bi-LSTM methods get lowest MAP score on this task. By analyzing the ranking list generated by Bi-LSTM model, we think this is mainly caused by three reasons. First, Bi-LSTM failed to consider the style constrain and many selected item are not consistent with condition style. Second, the images in Chuanda dataset is much more noisy. Unlike the clean images on Polyvore website, a lot of images in Chuanda contain irrelevant information such as: price tags, promotion ads etc. Third, Chuanda dataset is a smaller dataset than Polyvore. It is much more challenging to learn complex style concepts on relative small dataset with supervised learning methods.

In Figure \ref{fig:outfit-results}, we demonstrate the outfits generated with attributes pointwise mutual information ranking, Han et al's BiLSTM method, our HM-AIRL and fashion experts' selections for query tops under three distinct condition styles. Compared with experts' selection, only HM-AIRL guarantee both compatibility between items and consistency with condition style description. In the outfit generated by Bi-LSTM and PMI based algorithms, the selected  matching items actually belong to other styles.


In the experiment, we use Adam optimizer with a batch size of 256, learning rate of 0.00005 for the HM-ARIL optimization. For the MVAE model used to learn fashion item fusion representation, image encoder and decoder follow the standard DCGAN architecture \cite{radford2016unsupervised}. Attribute encoder and decoder is a standard 3-layer fully connected network VAE architecture. The two VAEs share the identical latent variables size of 256. The outfit style description encoder uses the base 12-layer BERT model that was pre-trained on Chinese Wikipedia corpus with 21128 unique Chinese characters, and we fix it during the training. The entire framework is implemented with Pytorch \cite{paszke2017automatic}.

\section{Remarks and Future Work}

Fashion experts keep proposing novel styles that are appreciated and imitated by individuals sharing the same taste or preference. In this work, we propose a framework to imitate fashion styles from outfit demonstrations. A hierarchical multimodal network is introduced to represent the whole outfit structure. Comparing with other work, our method captures the latent contextual information behind the fashion style by learning both the joint representation from image and attributes for each item and the compatibility and style consistency between items.

Relying on this hierarchical multimodal representation, we train the agent with an inverse reinforcement learning algorithm based on adversarial learning. Our approach builds upon a vast line of work on IRL. Hence, our approach, just like IRL, does not interact with the expert during training and adapt training samples to improve learning efficiency. Our experiment shows that HM-AIRL can learn the value function for imitating fashion styles and is robust to style shift. 

In recommendation, user behaviour data such as browsing history is very important. Content-based analysis such as style suggestion is only one factor. For the future, we like to integrate our framework into a full recommendation system and evaluate its performance. 
Note that the framework presented in this paper is not limited to fashion. Design artifacts in many domains contain latent concepts that can be expressed with sets of human-interpretable features capturing different levels of granularity \cite{Adar_2014_relation} \cite{Michailidou_2008_page}. This model also offers attractive capabilities: it can infer latent abstract concepts, and imitate experts from their demonstrations. In the future, we like to explore how this framework can empower applications in other domains such as interior design, architecture and etc.

{\small
\bibliographystyle{ieee_fullname}
\bibliography{egbib}
}

\end{document}